\documentclass{article} 

\usepackage{arxiv}
\usepackage{amsmath,amssymb,amsfonts}
\usepackage{bm}

\usepackage{subfigure}
\usepackage{multirow}
\usepackage{times}  
\usepackage{helvet}  
\usepackage{courier}  
\usepackage[hyphens]{url}  
\usepackage{graphicx} 
\usepackage{natbib}  
\usepackage{caption} 
\usepackage{xcolor}
\usepackage{enumitem}
\usepackage{booktabs}

\DeclareMathOperator{\expectation}{\mathbb{E}}

\usepackage{pifont}

\usepackage[ruled,linesnumbered,vlined]{algorithm2e}
\usepackage{nicefrac}

\usepackage{newfloat}
\usepackage{listings}
\DeclareCaptionStyle{ruled}{labelfont=normalfont,labelsep=colon,strut=off} 
\title{Multi-Agent Path Finding in Continuous Spaces with Projected Diffusion Models}
\author{ {\hspace{1mm}Jinhao Liang} \\
	Department of Computer Science\\
	University of Virginia\\
	Charlottesville, VA 22903, USA \\
	\texttt{jliang@email.virginia.edu} \\
	\And
	{\hspace{1mm}Jacob K. Christopher} \\
	Department of Computer Science\\
	University of Virginia\\
	Charlottesville, VA 22903, USA \\
	\texttt{csk4sr@virginia.edu} \\
	\AND
	{\hspace{1mm}Sven Koenig} \\
	Department of Computer Science\\
	University of California\\
	Irvine, CA 92697, USA \\
	\texttt{sven.koenig@uci.edu} \\
	\And
	{\hspace{1mm}Ferdinando Fioretto} \\
	Department of Computer Science\\
	University of Virginia\\
	Charlottesville, VA 22903, USA \\
	\texttt{fioretto@virginia.edu} \\
}

\begin{document}

\maketitle

\begin{abstract}
Multi-Agent Path Finding (MAPF) is a fundamental problem in robotics, requiring the computation of collision-free paths for multiple agents moving from their respective start to goal positions. Coordinating multiple agents in a shared environment poses significant challenges, especially in continuous spaces where traditional optimization algorithms struggle with scalability. Moreover, these algorithms often depend on discretized representations of the environment, which can be impractical in image-based or high-dimensional settings. Recently, diffusion models have shown promise in single-agent path planning, capturing complex trajectory distributions and generating smooth paths that navigate continuous, high-dimensional spaces. However, directly extending diffusion models to MAPF introduces new challenges since these models struggle to ensure constraint feasibility, such as inter-agent collision avoidance. To overcome this limitation, this work proposes a novel approach that integrates constrained optimization with diffusion models for MAPF in continuous spaces. This unique combination directly produces \emph{feasible} multi-agent trajectories that respect collision avoidance and kinematic constraints. The effectiveness of our approach is demonstrated across various challenging simulated scenarios of varying dimensionality.
\end{abstract}

\section{Introduction}

Multi-agent path finding (MAPF) is a critical problem in robotics and autonomous systems, where the goal is to compute collision-free paths for multiple agents navigating from their respective start positions to designated goals in a shared environment~\cite{stern2019multi}. 
This problem finds formulation in numerous domains, such as gaming, automated warehouses, and aircraft taxing~\cite{li2021lifelong}. 
The problem is inherently challenging due to the high-dimensional joint configuration space and the need for coordination among multiple agents to avoid collision. The complexity increases exponentially with the number of agents, making scalability a significant issue for traditional MAPF algorithms. Additionally, existing studies typically consider discrete environments ~\cite{stern2019multi,hopcroft1984complexity}, thus further limiting their applicability in scenarios in-the-wild.

The complexity of MAPF in continuous or high-dimensional environments calls for approaches that move beyond traditional discretized methods. Within this context, trajectory optimization has recently been tackled using diffusion models, a powerful class of generative models originally developed for tasks in image and signal processing~\cite{song2019generative,ho2020denoising}. These models approximate high-dimensional probability distributions by iteratively denoising sampled trajectories, leveraging strong inductive biases that provide effective heuristics even for very complex distributions. Their adaptability has accelerated their adoption across diverse engineering domains, including single-agent robotic path planning~\cite{carvalho2023motion,christopher2024constrained}. 
By learning the underlying distribution of (feasible) trajectories, diffusion models can produce diverse solutions that may be missed by traditional planners due to inductive bias. 
Additionally, these models possess the ability to generate smooth trajectories that effectively navigate high-dimensional spaces with complex obstacles while directly processing real-world representations of the environment. 

However, despite their potential, current diffusion models face significant challenges in generating \emph{feasible} trajectories. Existing approaches often rely on costly rejection sampling methods, which attempt to identify a feasible subset from a larger set of initially generated trajectories, if such a subset exists at all~\cite{carvalho2023motion,christopher2024constrained}. Additionally, despite their adoption in single-agent scenarios, extending diffusion models to MAPF presents additional challenges. The introduction of multiple agents requires the consideration of collision avoidance among agents, as well as kinematic constraints. 

{\em To address this limitation, this paper proposes a novel integration of constrained optimization in diffusion processes tailored to MAPF in continuous spaces.} The proposed method leverages the projection-based method for diffusion models~\cite{christopher2024constrained}, which has been recently introduced to steer the learned data distribution to satisfy some constraints of interest. 
This approach reformulates the sampling process as a constrained optimization problem, projecting the outputs of DMs at each sampling step into the feasible region. However, the MAPF feasible region is defined by a set of nonconvex nonlinear constraints (NNCs), which massively complicates the application of these diffusion models in scenarios with a large number of agents or with moving objects. To address this limitation and enhance computational efficiency, we propose an augmented Lagrangian method that relaxes the NNCs, making the proposed approach suitable for complex applications where classical MAPF algorithms fall short. This novel integration enables generative diffusion models to generate, for the first time, collision-free trajectories for scenarios involving dozens of agents and obstacles.

The paper's contributions are summarized as follows:
\begin{enumerate}[leftmargin=*, topsep=2pt, parsep=2pt]
    \item We introduce a novel formulation of MAPF in continuous spaces using diffusion models, enabling the simultaneous generation of trajectories for all agents in a single framework.
    \item To address the challenge of constraint satisfaction, we adapt the projection mechanism for MAPF by embedding constraints directly into the diffusion process, projecting the generated solutions into the feasible region.
    \item Given the computational intractability of MAPF in continuous spaces, especially with a large number of agents, we develop an augmented Lagrangian approach to accelerate the projection process. This enhancement significantly reduces computational overhead, making the method scalable and practical for real-world applications.
    \item We assess the ability of our approach to generate feasible MAPF trajectory empirically over several challenging scenarios, which include maps with narrow corridors, dense obstacles, and a large number of agents.
\end{enumerate}

\section{Related Work}
\paragraph{Multi-Agent Path Finding.}
The classical MAPF problem assumes that time and the environment are discretized into time steps and grids, respectively~\cite{stern2019multi}. Under this assumption, numerous search algorithms have been developed to efficiently obtain near-optimal solutions for MAPF in discrete environments, even for scenarios involving hundreds of agents~\cite{li2019improved, li2021eecbs,okumura2022priority, li2021eecbs}. While this assumption significantly reduces the complexity of MAPF, it creates a gap between the problem's formulation and real-world applications, posing challenges in many domains~\cite{shaoul2024multi}. Some studies attempt to generalize MAPF to continuous environments using probabilistic roadmaps~\cite{kavraki1996probabilistic} and rapidly exploring random trees~\cite{lavalle1998rapidly}. Another line of research formulates MAPF as a constrained optimization problem with continuous variables, employing methods such as sequential convex programming~\cite{6385823,7140034} and the alternating direction method of multipliers~\cite{10286111}. However, these methods often fail to find any solution if there are a large number of agents and obstacles, even if one exists.

\paragraph{Path Finding with Generative Models.}
There has been a growing interest in leveraging generative models for path finding problems. Existing studies primarily focus on using diffusion models to solve single-agent path finding problems~\cite{janner2022planning,carvalho2023motion}. Besides these approaches,~\citeauthor{okumura2022ctrms} utilizes a conditional variational autoencoder to predict cooperative timed roadmaps to aid in solving MAPF in continuous spaces.~\citeauthor{shaoul2024multi} uses diffusion models to generate a trajectory for a single agent and employs classical searching algorithms to determine the final solutions. However, these methods do not ensure the feasibility of the diffusion model outputs and cannot directly generate collision-free paths. In contrast, our approach integrates optimization techniques into diffusion models to directly generate feasible MAPF solutions in continuous spaces, even in scenarios with a significant number of obstacles, while ensuring feasibility.

\section{Preliminaries}
\paragraph{Diffusion Models.} Diffusion Models (DMs) are a class of probabilistic generative models designed to transform simple noise distributions into complex target data distributions. They operate through two Markov chains: (1) a forward diffusion process that progressively adds noise to data samples, and (2) a reverse denoising process that iteratively removes noise to recover data samples~\cite{yang2023diffusion}.
In the forward process, Gaussian noise is incrementally added to the data $\mathbf{x}_0 \sim q(\mathbf{x}_0)$ over $T$ timesteps, producing a sequence of noisy samples $\mathbf{x}_1, \mathbf{x}_2, \ldots, \mathbf{x}_T$: 
\begin{align}
\label{diff_forward} 
    q(\mathbf{x}_t | \mathbf{x}_{t-1}) = 
    \mathcal{N}\left(\mathbf{x}_{t}; \sqrt{1-\beta_t}\mathbf{x}_{t-1}, \beta_t\mathbf{I}\right), 
\end{align} 
    where $\beta_t \in (0,1)$ is a predefined variance schedule controlling the amount of noise added at each step, ensuring that the final distribution approximates an isotropic Gaussian.
The reverse process begins with a sample from the noise distribution $\mathbf{x}_T \sim \mathcal{N}(\mathbf{0}, \mathbf{I})$ and aims to reconstruct data samples by sequentially removing noise: 
\begin{align} \label{diff_reverse} 
    p_\theta(\mathbf{x}_{t-1} | \mathbf{x}_t) = 
    \mathcal{N}\left(\mathbf{x}_{t-1}; 
    \boldsymbol{\mu}_\theta(\mathbf{x}_t, t), \boldsymbol{\Sigma}_\theta(\mathbf{x}_t, t)\right), 
\end{align} 
    where $\theta$ represents the learned parameters of neural networks, and $\boldsymbol{\mu}_\theta$ and $\boldsymbol{\Sigma}_\theta$ are functions parameterizing the mean and covariance, respectively. Through this process, DMs iteratively transform random noise samples into data resembling the target distribution $q(\mathbf{x}_0)$. 

\textbf{Score-based DMs} employ a neural network $\boldsymbol{s}_\theta$ to approximate the score function $\nabla{\mathbf{x}_t} \log q(\mathbf{x}_t)$, which points in the direction of the steepest ascent of the data density at each noise level~\cite{song2020score}. 
The training objective is to minimize the difference between the true score and the network's approximation~\cite{yang2023diffusion}: 
\begin{subequations} \begin{align} 
&{\expectation}_{\substack{ 
    t \sim {1,\dots,T}, \ 
    \mathbf{x}_0 \sim q(\mathbf{x}_0), \\ 
    \mathbf{x}_t \sim q(\mathbf{x}_t | \mathbf{x}_0) }} 
    \left| \nabla{\mathbf{x}_t} \log q(\mathbf{x}_t) - \boldsymbol{s}_\theta(\mathbf{x}_t, t) \right|^2 \eta(t) \beta_t \\ 
    ={}&{\expectation}_{\substack{ 
    t \sim {1,\dots,T}, \ \mathbf{x}_0 \sim q(\mathbf{x}_0), \notag \\
    \mathbf{x}_t \sim q(\mathbf{x}_t | \mathbf{x}0) }} 
    | \nabla{\mathbf{x}_t} \log q(\mathbf{x}_t | \mathbf{x}_0) -  \boldsymbol{s}_\theta(\mathbf{x}_t, t) |^2 \eta(t) \beta_t + \text{const}, 
    \end{align} 
\end{subequations} 
    where 
    $q(\mathbf{x}_t | \mathbf{x}_0) =
    \mathcal{N}\left(\mathbf{x}_{t}; \sqrt{1-\beta_t}\mathbf{x}_{0}, \beta_t\mathbf{I}\right)$ and $\eta(t)$ is a positive weighting function.

As shown by~\citet{yang2023diffusion}, classical DMs are a special case of score-based DMs. In the subsequent sections, our focus will be on score-based DMs due to their flexibility and effectiveness.

\subsection{Multi-Agent Path Finding in Continuous Space}
Multi-Agent Path Finding (MAPF) involves computing collision-free trajectories for multiple agents moving from their respective start locations to designated goals within a shared environment. Consider a set of $N_a$ agents $\mathcal{A} = \{ a_1, a_2, \ldots, a_{N_a} \}$ operating on a two-dimensional plane, in a continuous space. 
Each agent $a_i$ is modeled as a sphere with radius $r_i$ and has a trajectory over $H$ time steps denoted by $\boldsymbol{\pi}_i = [\pi_i^1, \pi_i^2, \ldots, \pi_i^H]$, where $\pi_i^h = (x_i^h, y_i^h)$ represents the position of agent $a_i$ at time $h$. 
The agents have start positions $\mathbf{B} = [b_1, b_2, \ldots, b_{N_a}]$ and goal positions $\mathbf{E} = [e_1, e_2, \ldots, e_{N_a}]$. 
In addition, their movement must adhere to kinematic constraints, such as maximum velocities.
The environment contains $N_o$ obstacles $\mathcal{O} = \{ o_1, \dots, o_{N_o} \}$. 
 The objective is to find a set of trajectories $\boldsymbol{\Pi} = \{\boldsymbol{\pi}_1, \boldsymbol{\pi}_2, \ldots, \boldsymbol{\pi}_{N_a}\}$, each associated with agent $a_i$, that minimizes a cost function while ensuring feasibility with respect to environmental constraints and inter-agent collision avoidance. 
 
The MAPF problem can be formulated as the following constrained optimization: 
\begin{subequations} 
    \begin{align} 
    \min_{\boldsymbol{\Pi}} & \quad \mathcal{J}(\boldsymbol{\Pi}) \label{mapf_objective} \\ 
    \text{s.t.} & \quad \boldsymbol{\Pi} \subseteq \Omega_{\text{obs}}, \label{mapf_constraint_env} \\ 
    & \quad \pi_i^1 = b_i, \quad \forall i \in [N_a], \label{mapf_constraint_start} \\ 
    & \quad \pi_i^H = e_i, \quad \forall i \in [N_a], \label{mapf_constraint_goal} \\
    & \quad \text{Kinematic constraints on } \boldsymbol{\Pi}, \label{mapf_constraint_kinematic} \\ 
    & \quad \text{Collision avoidance between agents in } \boldsymbol{\Pi}, \label{mapf_constraint_collision} 
    \end{align} 
\end{subequations} 
    where $\mathcal{J}: \mathbb{R}^{N_a \times H \times 2} \rightarrow \mathbb{R}$ is the cost function (e.g., total travel time or energy consumption), and $\Omega_{\text{obs}}$ denotes the feasible region of the environment considering obstacles. 
    Constraints \eqref{mapf_constraint_env} ensure that agents avoid obstacles, 
    \eqref{mapf_constraint_start} and \eqref{mapf_constraint_goal} ensure that each agent starts at and reaches its designated start and end positions, respectively,
    \eqref{mapf_constraint_kinematic} enforce kinematic limits, and \eqref{mapf_constraint_collision} prevent inter-agent collisions. In the following, we denote the constraint set \eqref{mapf_constraint_start}-- \eqref{mapf_constraint_collision}, with $\Omega$.
    
The MAPF problem is challenging due to the high dimensionality of the joint configuration space and the need to coordinate multiple agents simultaneously~\cite{stern2019multi,shaoul2024multi}. Traditional methods often struggle with scalability and may not efficiently handle the continuous nature of real-world environments, as shown in~\cite{6385823,shaoul2024multi}. We seek to address this issue using constrained DMs.

\section{Constrained Diffusion Models}
In this section, we first revisit the sampling process for DMs and then investigate the integration of DMs and optimization to constrain the output of DMs satisfying constraints.

\subsection{Recall The Sampling Process in DMs}
Since the sampling process in DMs is a Markov process, we generate $\mathbf{x}_{0}$ by iterative sampling from the conditional distribution $\mathbf{x}_{t} \sim q(\mathbf{x}_{t} | \mathbf{x}_0)$ as $t \rightarrow 0$, where $q(\mathbf{x}_t | \mathbf{x}_0)$ shifts from Gaussian noise to the training data distribution as $t$ decreases. 
The sample is optimized with respect to each interim data distribution by $M$ iterations of Stochastic Gradient Langevin Dynamics (SGLD):
\begin{equation}
    \label{eq:sgld}
    \mathbf{x}_{t}^{i+1} = \mathbf{x}_{t}^{i} + \gamma_t \nabla_{\mathbf{x}_{t}^{i}} \log q(\mathbf{x}_{t}^i|\mathbf{x}_0) + \sqrt{2\gamma_t}\mathbf{z},
\end{equation}
where $\mathbf{z}$ is standard normal,  $\gamma_t > 0$ is the step size, and 
$\nabla_{\mathbf{x}_{t}^{i}} \log q(\mathbf{x}_{t}^i|\mathbf{x}_0)$ is approximated by the learned score function $\boldsymbol{s}_{\theta}(\mathbf{x}_t, t)$.

\citet{christopher2024constrained} derive theory connecting the application of SGLD for sampling to iterative, gradient-based optimization algorithms. The described process ensures that, under appropriate conditions, samples are distributed according to the target distribution \( q(\mathbf{x}_t) \). 
As shown by~\citet{christopher2024constrained}, SGLD converges toward a stationary distribution under mild assumptions, transitioning toward deterministic gradient ascent as the stochastic component diminishes. This connects the reverse diffusion process to an optimization problem, minimizing the negative log-likelihood of the data distribution and forming the foundation for constrained sampling via iterative projections.

\begin{algorithm}[t]
\DontPrintSemicolon
\caption{PDM}
\label{alg:pgd_annealed_ld}
{
$\mathbf{x}_T^0 \sim \mathcal{N}(\mathbf{0}, \sqrt{\beta_T} \mathbf{I})$\\
\For{$t = T$ \KwTo $1$}{
    $\gamma_t \gets \nicefrac{\beta_{t}}{2 \beta_{T}}$\\
    \For{$i = 1$ \KwTo $M$}{
        $\mathbf{z} \sim \mathcal{N}(\mathbf{0}, \mathbf{I})$; \,\,
        $\mathbf{g} \gets \boldsymbol{s}_{\theta^*}(\mathbf{x}_t^{i-1}, t)$\\
        $\mathbf{x}_{t}^{i} = \mathcal{P}_{\Omega}(\mathbf{x}_{t}^{i-1} + \gamma_t \mathbf{g} + \sqrt{2\gamma_t}\mathbf{z})$\;
    }
    $\mathbf{x}_{t-1}^0 \gets \mathbf{x}_t^M$\;
}
\Return $\mathbf{x}_0^0$\;
}
\end{algorithm}
\subsection{Projected Diffusion Models}

In this subsection, we introduce Projected Diffusion Models (PDM) to ensure that generated outputs satisfy predefined constraints. While the objective remains consistent with traditional score-based DMs, the solution is restricted to lie within a feasible region \(\Omega\). This transforms the optimization problem into a constrained formulation~\cite{christopher2024constrained}:
\begin{subequations}
\label{eq:constrained_optimization}
    \begin{align}
        \label{eq:constrained-diffusion}
        \min_{\mathbf{x}_{T}, \ldots, \mathbf{x}_{1}} &\ \sum_{t = T, \ldots, 1} - \log q(\mathbf{x}_{t}|\mathbf{x}_0) \\
        \label{eq:constrained-diffusion-constr}
        \textrm{s.t.}  &\quad \mathbf{x}_{T}, \ldots, \mathbf{x}_{0} \in \Omega.
    \end{align}
\end{subequations}
The reverse sampling process in PDM aligns closely with that of traditional score-based DMs. Specifically, the score network \(\boldsymbol{s}_{\theta}(\mathbf{x}_{t}, t)\) estimates the gradient of the objective in Equation~\eqref{eq:constrained-diffusion}, enabling iterative updates as defined in Equation~\eqref{eq:sgld}. However, the presence of constraints \eqref{eq:constrained-diffusion-constr} necessitates a modification to the update rule to maintain feasibility. To address this, PDM employs a projected guidance approach, incorporating constraints into the optimization process.

The projection operator, \(\mathcal{P}_{\Omega}\), is defined as solving a constrained optimization problem:
\begin{equation}
    \label{eq:projection}
    \mathcal{P}_{\Omega}(\mathbf{x}) = \arg \min_{\mathbf{y} \in \Omega} d(\mathbf{x},\mathbf{y}),
\end{equation}
where $d(\mathbf{x},\mathbf{y})$ is a distance function, and, unless otherwise $d(\mathbf{x},\mathbf{y})$ denotes the euclidean distance: $\|\mathbf{y} - \mathbf{x}\|_{2}^2$, which identifies the closest feasible point \(\mathbf{y}\) within \(\Omega\) to the input \(\mathbf{x}\). 

To ensure feasibility at each step, the projected diffusion model applies the projection operator after updating \(\mathbf{x}_{t}\), leading to the \emph{projected diffusion model sampling} step:
\begin{equation}
    \label{eq:reverse-pgd}
    \mathbf{x}_{t}^{i+1} = \mathcal{P}_{\Omega} \left(\mathbf{x}_{t}^{i} + \gamma_t \nabla_{\mathbf{x}_{t}^{i}} \log q(\mathbf{x}_{t}|\mathbf{x}_0) + \sqrt{2\gamma_t}\mathbf{z}\right),
\end{equation}
where $\Omega$ is the set of constraints and $\mathcal{P}_{\Omega}$ is a projection onto $\Omega$. 
Throughout the reverse Markov chain, each iteration performs a gradient step to minimize the objective in Equation~\eqref{eq:constrained-diffusion}, while ensuring feasibility through projection. As is the case in this paper, the complete sampling process is outlined in Algorithm~\ref{alg:pgd_annealed_ld}.

PDM directly minimizes the negative log-likelihood as its core objective, similar to standard unconstrained sampling methods. This approach provides a crucial benefit: \emph{it directly optimizes the probability of generating samples that align with the data distribution}, while simultaneously imposing explicit, verifiable constraints. In the next section, we develop a projection mechanism tailored for MAPF in continuous spaces.

\section{Efficient Projections for MAPFs}
While PDM provides a useful method to steer samples generated by the generative model to satisfy relevant constraints, projecting onto nonconvex sets can be a computationally expensive operation, especially when it is required to be computed at each step of the sampling process.
To address this shortcoming, we develop a projection mechanism to generate feasible trajectories for all agents. To accelerate the projection process, we adopt the augmented Lagrangian method (ALM)~\cite{boyd2011distributed} to the projection process.

\subsection{Collision-free Trajectories Projection Mechanism}
In the following, we define the mathematical formulation of the feasible region $\Omega$ for the MAPF problem, distinguishing between convex and nonconvex constraints.

\paragraph{Convex Constraints.}
First, each agent's trajectory must start and end its specified start and goal points, as specified in Constraints \eqref{mapf_constraint_start} and \eqref{mapf_constraint_goal}.

Additionally, agents must adhere to maximum velocity limits between consecutive time steps: 
\begin{align} 
\left( \pi_i^{h} - \pi_i^{h-1} \right)^2 \leq 
    \left( v_i^\text{max} \Delta t \right)^2, 
    \quad \forall i \in [N_a], h \in \{2, \dots, H\}, \label{eq:convex_velocity} 
\end{align} 
    where $v_i^\text{max}$ denotes the maximum allowable velocity for agent $a_i$, and $\Delta t$ is the time interval between steps.

Together, these constraints define a convex set: 
\begin{align} 
    \Omega_c = 
    \left\{ \boldsymbol{\Pi} \in \mathbb{R}^{N_a \times H \times 2} \Big| \text{Constr. } \eqref{mapf_constraint_start}, \eqref{mapf_constraint_goal}, \text{ and } \eqref{eq:convex_velocity} \text{ hold} \right\}. 
\end{align}

\paragraph{Nonconvex Constraints.}
To ensure collision avoidance between agents, we impose the following nonconvex constraints: 
\begin{align}
    (\pi_i^h - \pi_j^{h})^2 \geq (R^a)^2, \forall i, j, \ i \neq j \in [N_a], h \in [H], \label{nonconvex_agent_collision}
\end{align}
where $R^a$ denotes the minimum distance between agents at each time. 

Similarly, to avoid collisions between agents and static obstacles, we have: 
\begin{align}
    (\pi_i^h - o_j)^2 \geq (R^o)^2, \forall i, j \in [N_a], h \in [H], \label{nonconvex_obstacle_collision}
\end{align}
where $R^o$ denotes the minimum distance between agents and obstacles to guarantee noncollision. Similarly, these two constraints define 
\begin{align} 
    \Omega_n = 
    \left\{ \boldsymbol{\Pi} \in \mathbb{R}^{N_a \times H \times 2} \Big| \text{Constr. } \eqref{nonconvex_agent_collision},  \eqref{nonconvex_obstacle_collision}, \text{ hold} \right\}. 
\end{align}

The complete feasible set is given by: $\Omega = \Omega_c \cap \Omega_n$. Although the projector $\mathcal{P}_\Omega$ can generate feasible MAPF trajectories, the nonconvex constraints result in high computational costs.

\subsection{ALM for Efficient Projection}
To address this issue, we seek to relax the nonconvex constraints in MAPF to transform the original nonconvex quadratically constrained quadratic problem (QCQP) into a convex QCQP. To facilitate analysis, we rewrite the inequality constraints as equalities:
\begin{subequations}
\begin{align}
     \mathcal{H}_a: &(R^a)^2 - (\pi_i^h - \pi_j^{h})^2 + d_{i,j,h}^a = 0, \forall i, j, \ i \neq j, \ \forall h, \label{equality_agent_collision} \\
    \mathcal{H}_o:&(R^o)^2 - (\pi_i^h - o_j)^2 + d_{i,j,h}^o = 0, \forall i, j, \ \forall h, \label{equality_obstacle_collision}  
\end{align}
\end{subequations}
where $d_{i,j,h}^a$ and $d_{i,j,h}^o$ (with vector form $\boldsymbol{d^a}$ and $\boldsymbol{d^o}$, respectively) are positive dummy variables. The Lagrangian function is defined as:
\begin{align}
    \mathcal{L}_\text{c}(\boldsymbol{\Pi}, \boldsymbol{\nu}_a, \boldsymbol{\nu}_o) = f(x) + \boldsymbol{\nu}_a^\top \mathcal{H}_a(\boldsymbol{\Pi}) + \boldsymbol{\nu}_o^\top \mathcal{H}_o(\boldsymbol{\Pi}),
\end{align}
where $\boldsymbol{\nu}_a$ and $\boldsymbol{\nu}_o$ are Lagrangian multipliers, $\mathcal{H}_a$ and $\mathcal{H}_o$ represent the equality constraints defined by \eqref{equality_agent_collision} and \eqref{equality_obstacle_collision}, respectively. Specifically, $\mathcal{H}_a$ corresponds to the agent collision avoidance constraints $(R^a)^2 - (\pi_i^h - \pi_j^{h})^2 + d_{i,j,h}^a = 0, \forall i, j, \ i \neq j, \ \forall h,$ and $\mathcal{H}_o$ corresponds to the obstacle collision avoidance constraints $(R^o)^2 - (\pi_i^h - o_j)^2 + d_{i,j,h}^o = 0, \forall i, j, \ \forall h$. To improve the poor convergence of the classical lagrangian function, we can augment the Lagrangian function with a penalty on the constraint residuals~\cite{boyd2011distributed,kotary2022fast}:
\begin{equation}
\begin{aligned}
    \mathcal{L}(\boldsymbol{\Pi}, \boldsymbol{\nu}_a, \boldsymbol{\nu}_o) = f(x) + & \boldsymbol{\nu}_a^\top \mathcal{H}_a(\boldsymbol{\Pi}) + \boldsymbol{\nu}_o^\top \mathcal{H}_0(\boldsymbol{\Pi}) + \rho_a \| \mathcal{H}_a(\boldsymbol{\Pi}) \|^2 + \rho_o \| \mathcal{H}_o(\boldsymbol{\Pi}) \|^2,
\end{aligned}
\end{equation}
where $\rho_a$ and $\rho_o$ are chosen penalty weights on the equality residuals. The corresponding Lagrangian Dual function can be defined:
\begin{align}
    \boldsymbol{d}(\boldsymbol{\nu}_a, \boldsymbol{\nu}_o) = \min_{\boldsymbol{\Pi}} \mathcal{L}(\boldsymbol{\Pi}, \boldsymbol{\nu}_a, \boldsymbol{\nu}_o).
\end{align}

The Lagrangian Dual Problem is to maximize the dual function:
\begin{subequations}
\label{dual_problem}
\begin{align}
    \arg \max_{\boldsymbol{\nu}_a, \boldsymbol{\nu}_o} &\quad\boldsymbol{d}(\boldsymbol{\nu}_a, \boldsymbol{\nu}_o) \\
    \text{s.t.}  &\quad \boldsymbol{\Pi} \in \Omega_c.
\end{align}   
\end{subequations}

Through weak duality, solving the dual problem (\ref{dual_problem}) can provide a lower bound for the original problem's optimal solution. Specifically, a feasible solution $\hat{\boldsymbol{\Pi}}$ to the Primal problem can be derived from the dual solution ($\boldsymbol{\nu}_a^*, \boldsymbol{\nu}_o^*$) via the stationarity condition:
\begin{subequations}
\label{primal_problem}
\begin{align}
    \hat{\boldsymbol{\Pi}} = \arg \min_{\boldsymbol{\Pi} \in \Omega_c} &\quad \mathcal{L}(\boldsymbol{\Pi}, \boldsymbol{\nu}_a^*, \boldsymbol{\nu}_o^*).
\end{align}   
\end{subequations}

The dual problem (\ref{dual_problem}) can be solved iteratively, named the Dual Ascent method (DAM):
\begin{subequations}
\begin{align}
    \boldsymbol{\Pi}^k = \ & \arg \min_{\boldsymbol{\Pi} \in \Omega_c} \ \mathcal{L}(\boldsymbol{\Pi}, \boldsymbol{\nu}_a^k, \boldsymbol{\nu}_o^k), \\
    \boldsymbol{\nu}_a^{k+1} = \ & \boldsymbol{\nu}_a^{k} + \rho_a^k \mathcal{H}_a(\boldsymbol{\Pi}^k), \\   
    \boldsymbol{\nu}_o^{k+1} = \ & \boldsymbol{\nu}_a^{k} + \rho_o^k \mathcal{H}_o(\boldsymbol{\Pi}^k).
\end{align}
\end{subequations}

Using ALM significantly accelerates the projection process, especially in complex scenarios. The augmented sampling process is described in Algorithm~\ref{alg:alm_pgd_annealed_ld}.

\begin{algorithm}[t]
\caption{ALM for Projection}
\label{alg:alm_pgd_annealed_ld}
{
\KwIn{Tolerance $\delta$, Weight $\rho$, Initial Trajectory $\hat{\boldsymbol{\Pi}}$}
$\mathbf{x}_T^0 \sim \mathcal{N}(\mathbf{0}, \sigma_T \mathbf{I})$\\
\While{$\nabla_{\boldsymbol{\nu}_a} < \delta 
\wedge \nabla_{\boldsymbol{\nu}_o} < \delta$}{
    $\hat{\boldsymbol{\nu}}_a \gets \mathcal{H}_a(\hat{\boldsymbol{\Pi}})$, $\hat{\boldsymbol{\nu}}_o \gets \mathcal{H}_o(\hat{\boldsymbol{\Pi}})$\;
    $\hat{\boldsymbol{\Pi}} \gets  \arg \min_{\boldsymbol{\Pi} \in \Omega_c} \ \mathcal{L}(\hat{\boldsymbol{\Pi}}, \boldsymbol{\nu}_a^*, \boldsymbol{\nu}_o^*)$\;
    $\nabla_{\boldsymbol{\nu}_a} \gets \mathcal{H}_a(\hat{\boldsymbol{\Pi}})$, $\nabla_{\boldsymbol{\nu}_o} \gets \mathcal{H}_o(\hat{\boldsymbol{\Pi}})$ \;
    $\rho \gets \text{Update}(\rho)$ 
}
\Return $\hat{\boldsymbol{\Pi}}$\;
}
\end{algorithm}

\section{Experiments}
We evaluate the performance of PDM in generating feasible trajectories for MAPF in continuous spaces. We compare PDM against standard Diffusion Models (SDM) and Guided Diffusion Models (GDM) across three challenging scenarios: Narrow Corridors, Obstacle-Dense Environments, and Agent-Dense Environments.

\subsection{Experimental Setup}

We conduct experiments in the following scenarios:

\begin{itemize} \item \textbf{Narrow Corridors}: Scenarios where agents must exchange positions in confined spaces, requiring precise coordination to avoid collisions. \item \textbf{Obstacle-Dense Environments}: Scenarios with a high density of obstacles, where agents must navigate complex paths to reach their goals without collisions. \item \textbf{Agent-Dense Environments}: Scenarios with a large number of agents, increasing the complexity of collision avoidance and coordination. \end{itemize}

For each scenario, we generate environments where the positions of obstacles and agents are randomly assigned and do not appear in the training data, ensuring that the models are tested on unseen configurations. The training data is collected following the routine described in~\cite{okumura2022ctrms}.

We evaluate the methods based on two metrics:

\begin{itemize} \item \textbf{Violation Rate}: The percentage of constraints violated, indicating the feasibility of the generated trajectories. \item \textbf{Total Path Length}: The sum of the lengths of the paths taken by all agents, reflecting the efficiency of the trajectories. \end{itemize}

We compare our proposed PDM against the following baseline methods:

\begin{itemize} \item \textbf{Standard Diffusion Models (SDM)}: Standard diffusion models used to generate trajectories without any constraint handling. \item \textbf{Guided Diffusion Models (GDM)}: Diffusion models guided by penalty terms added during the sampling process to encourage feasibility, similar to the method used in~\cite{carvalho2023motion}. \end{itemize}

\subsection{Evaluation on Narrow Corridors}
\begin{figure*}[b]
    \centering
    \subfigure[Narrow Corridor 1.]
    {
        \includegraphics[width=0.97\linewidth]{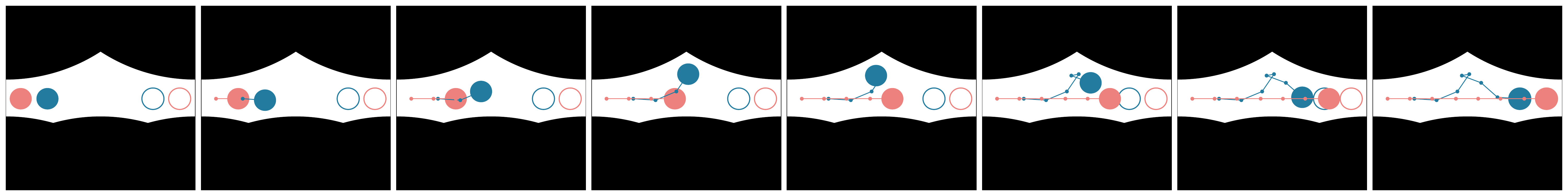}
        \label{fig:narrow_corridor1}
    }
    \subfigure[Narrow Corridor 2.]
    {
        \includegraphics[width=0.97\linewidth]{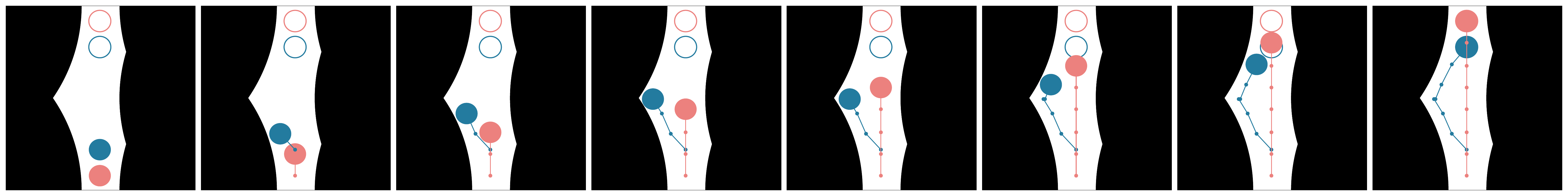}
        \label{fig:narrow_corridor2}
    }
    \caption{Collision-free trajectories generated by PDM in Narrow Corridor scenarios. Agents (solid circles) navigate to their goals (empty circles) by exchanging positions in confined spaces without collisions.}
    \label{fig:narrow_corridors}
\end{figure*}
The Narrow Corridor scenarios are designed to test the ability of the methods to generate feasible trajectories in tight spaces where agents must exchange positions. Figures~\ref{fig:narrow_corridor1} and \ref{fig:narrow_corridor2} illustrate the trajectories generated by PDM in two different narrow corridor scenarios. Agents (solid circles) successfully reach their respective goals (empty circles) by coordinating their movements to avoid collisions in the confined space.
Notice how PDM can identify a set of feasible paths for each agent in the narrow corridor by generating complex maneuvers and adjusting speed and position to allow an agent to overtake the other.

Table~\ref{tab:narrow_corridors} presents the performance of all methods in terms of violation rate and total path length for the two narrow corridor scenarios (NC1 and NC2). Lower values indicate better performance.
Notice how PDM outperforms both DM and GDM, \emph{achieving zero violation rates} and the \emph{shortest total path lengths in both scenarios}. In contrast, standard DM exhibits high violation rates and longer paths, indicating significant limitations in handling constraints. GDM reduces violation rates compared to DM but still falls short of PDM's performance.

\begin{table}[t]
\renewcommand{\arraystretch}{1.3}
\centering
\resizebox{0.4\columnwidth}{!}{
\begin{tabular}{ccccc}
\toprule
&      & PDM    & DM     & GDM    \\ 
\midrule
\multirow{2}{*}{\begin{tabular}[c]{@{}c@{}}Violation\\ Rate\end{tabular}} & NC 1 & 0      & 34.62  & 0.96   \\
& NC 2 & 0      & 15.79  & 5.26   \\ \hline
\multirow{2}{*}{\begin{tabular}[c]{@{}c@{}}Path\\ Length\end{tabular}}    & NC 1 & 0.7867 & 2.6766 & 0.8235 \\
 & NC 2 & 0.7521 & 2.1293 & 0.8398 \\ \bottomrule
\end{tabular}}
\caption{Performance Evaluation on Narrow Corridors.}
\label{tab:narrow_corridors}
\end{table}

\subsection{Evaluation on Obstacle-dense Scenarios}
\begin{figure}[t]
    \centering
    \subfigure[Obstacle-dense Scenario 1.]
    {
        \includegraphics[width=0.3\columnwidth]{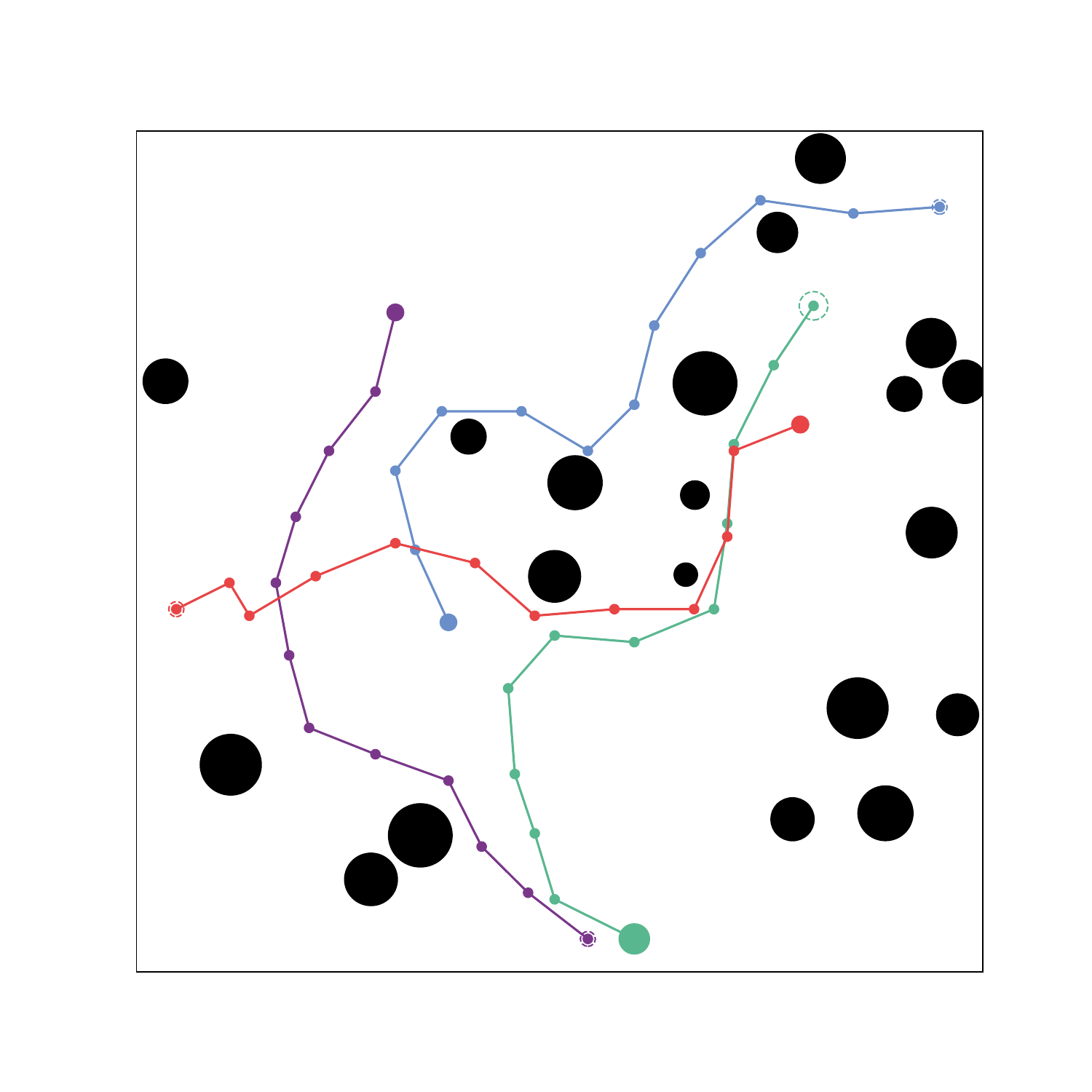}
        \label{fig:obstacle_dense1}
    }
    \subfigure[Obstacle-dense Scenario 2.]
    {
        \includegraphics[width=0.3\columnwidth]{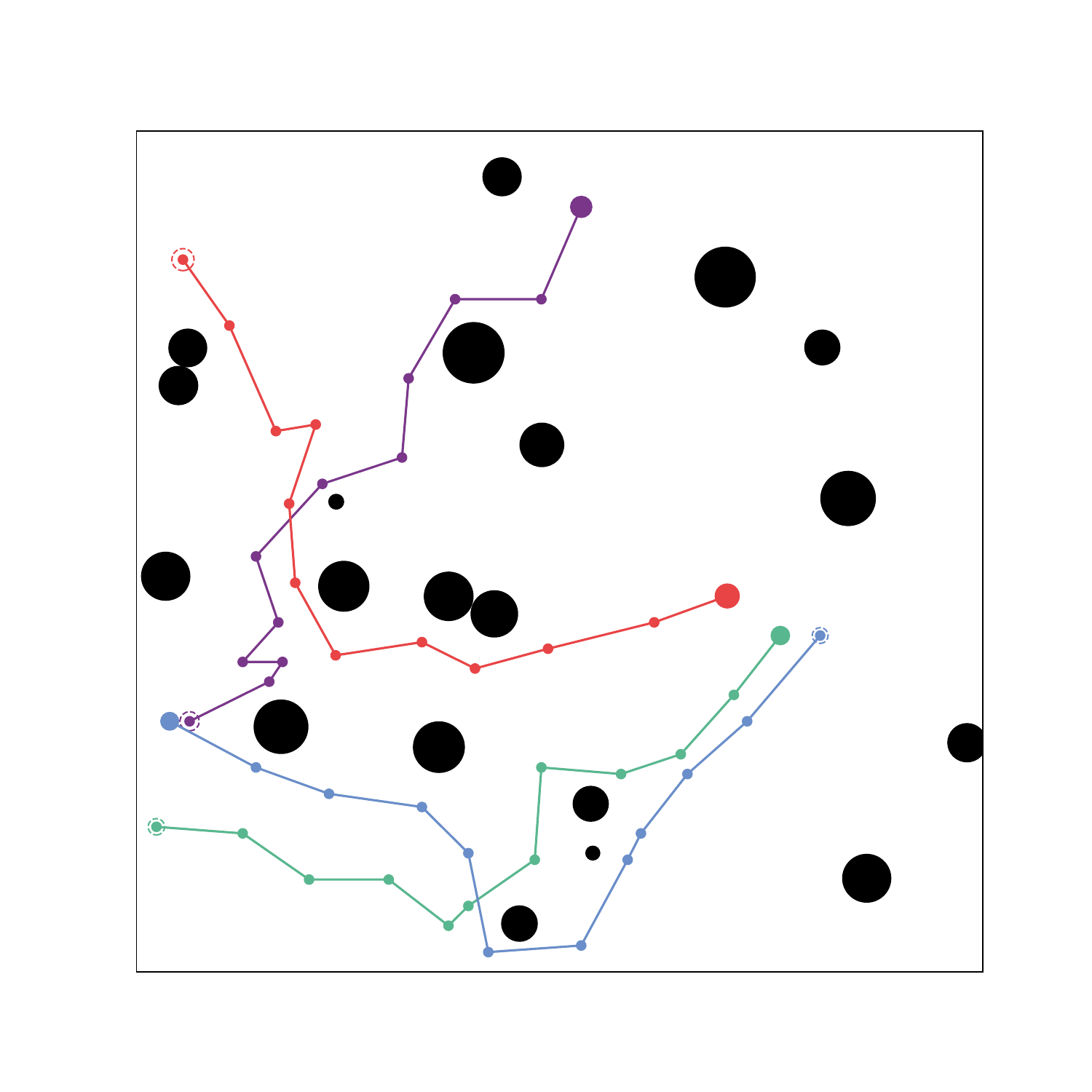}
        \label{fig:obstacle_dense2}
    }
    \caption{Collision-free trajectories generated by PDM in Obstacle-Dense scenarios. Agents successfully navigate through environments with numerous obstacles to reach their goals. The empty dashed circles denote starting points, and the solid circles represent the goals.}
    \label{Obstacle-dense Scenarios.}
\end{figure}

In the Obstacle-Dense scenarios, we test the methods in environments with twenty randomly placed obstacles and four agents. Figures~\ref{fig:obstacle_dense1} and \ref{fig:obstacle_dense2} show the trajectories generated by PDM, demonstrating its ability to navigate complex environments while avoiding collisions even when agents need to navigate scenarios presenting a large number of obstacles. 

\begin{table}[b]
\centering
\renewcommand{\arraystretch}{1.3}
\resizebox{0.4\columnwidth}{!}{
\begin{tabular}{ccccc}
\toprule
&      & PDM    & DM     & GDM    \\ \midrule
\multirow{2}{*}{\begin{tabular}[c]{@{}c@{}}Violation\\ Rate\end{tabular}} & OS 1 & 0      & 0.58   & 0.48   \\
& OS 2 & 0      & 1.02    & 0.58   \\ \hline
\multirow{2}{*}{\begin{tabular}[c]{@{}c@{}}Path\\ Length\end{tabular}}    & OS 1 & 2.0087 & 6.0228 & 5.6979 \\
& OS 2 & 2.0457 & 5.8585 & 5.2771 \\ \bottomrule
\end{tabular}}
\caption{Performance Evaluation on Obstacle-dense Scenarios.}
\label{tab:obstacle_dense}
\end{table}

Table~\ref{tab:obstacle_dense} summarizes the performance of the methods in the obstacle-dense scenarios (OD1 and OD2).
Notice how PDM achieves the best performance, maintaining zero violation rates and the shortest paths, indicating strong adaptability to dense obstacles. In contrast, DM exhibits higher violation rates and the longest paths. GDM outperforms DM but is worse than PDM. These results emphasize PDM's robustness and efficiency.

\begin{table}[t]
\centering
\renewcommand{\arraystretch}{1.3}
\resizebox{0.4\columnwidth}{!}{
\begin{tabular}{ccccc}
\toprule
&      & PDM    & DM      & GDM    \\ \midrule
\multirow{2}{*}{\begin{tabular}[c]{@{}c@{}}Violation\\ Rate\end{tabular}} & AS 1 & 0.31   & 3.78    & 0.54   \\
 & AS 2 & 0.17   & 2.99    & 0.37   \\ \hline
\multirow{2}{*}{\begin{tabular}[c]{@{}c@{}}Path\\ Length\end{tabular}}    & AS 1 & 5.2021 & 11.1599 & 5.4932 \\
& AS 2 & 5.1631 & 11.4114 & 5.4081 \\ \bottomrule
\end{tabular}}
\caption{Performance Evaluation on Agent-dense Scenarios.}
\label{Performance Evaluation on Agent-dense Scenarios.}
\end{table}

\subsection{Evaluation on Agent-dense Scenarios}
Finally, we test the ability of our proposed method to handle a large collection of agents. An increasing number of agents significantly introduces computational costs during projection, which makes standard projection methods challenging to handle. To address this, we use the ALM method to efficiently address Agent-dense Scenarios. Table \ref{Performance Evaluation on Agent-dense Scenarios.} evaluates PDM, DM, and GDM in agent-dense scenarios (AS 1 and AS 2). PDM achieves the lowest violation rates (0.31 and 0.17) and shortest path lengths (5.2021 and 5.1631), highlighting its efficiency in handling high agent density. GDM also shows moderate performance, with higher violation rates and longer path lengths compared to PDM. DM performs the worst, with significantly higher violation rates (3.78 and 2.99) and longest paths (11.1599 and 11.4114), indicating limited suitability for agent-dense conditions.

{\em These results are significant as they demonstrate the power of combining diffusion models with constrained optimization techniques to address problems that would be otherwise challenging to be tackled by these two areas independently.}

\section{Conclusion}
In this paper, we have presented a novel approach that combines constrained optimization techniques with DMs to generate collision-free trajectories for MAPF in continuous spaces. By integrating constraints directly into the diffusion process, our method enables the direct generation of feasible solutions for MAPF without the need for expensive rejection sampling or post-processing steps. This integration ensures that the generated trajectories satisfy all necessary constraints, including collision avoidance between agents, adherence to kinematic limits, and compliance with start and goal positions.

To address the computational challenges inherent in handling complex constraints, especially in scenarios with a large number of agents or obstacles, we designed an ALM to efficiently manage the projection process within the diffusion framework. The ALM significantly accelerates the computation by transforming the constrained optimization problem into a series of unconstrained problems augmented with penalty terms and Lagrange multipliers. This enhancement makes our approach scalable and practical for real-world applications where computational resources and time are critical factors.

Our preliminary experiments across various challenging scenarios—including narrow corridors, obstacle-dense environments, and agent-dense environments—demonstrate the effectiveness and robustness of our proposed method. In narrow corridor scenarios, where precise coordination is crucial, our PDM successfully generated feasible trajectories that allowed agents to exchange positions without collisions. In obstacle-dense environments, PDM consistently navigated agents through complex paths while maintaining zero violation rates and optimizing path lengths. In agent-dense scenarios, despite the increased complexity due to the higher number of agents, PDM maintained superior performance with the lowest violation rates and shortest total path lengths.

Crucially, the integration of constrained optimization into the diffusion process not only ensures constraint satisfaction but also improves the overall quality of the generated trajectories. By embedding constraint handling directly within the generative model, the aim is to eliminate the reliance on heuristic adjustments, leading to a cohesive and effective solution. We hope that applications like this one would help to bridge the gap between probabilistic generative models and constrained optimization, opening new avenues for applying diffusion models to complex multi-agent robotic systems.

\section*{Acknowledgment}
This research is partially supported by NSF grants 2334936, 2334448, and NSF CAREER Award 2401285. 
The authors acknowledge Research Computing at the University of Virginia for providing computational resources that have contributed to the results reported within this paper. 
The views and conclusions of this work are those of the authors only.

\bibliographystyle{unsrtnat}
\bibliography{references}

\begin{thebibliography}{22}
\providecommand{\natexlab}[1]{#1}
\providecommand{\url}[1]{\texttt{#1}}
\expandafter\ifx\csname urlstyle\endcsname\relax
  \providecommand{\doi}[1]{doi: #1}\else
  \providecommand{\doi}{doi: \begingroup \urlstyle{rm}\Url}\fi

\bibitem[Stern et~al.(2019)Stern, Sturtevant, Felner, Koenig, Ma, Walker, Li, Atzmon, Cohen, Kumar, et~al.]{stern2019multi}
Roni Stern, Nathan Sturtevant, Ariel Felner, Sven Koenig, Hang Ma, Thayne Walker, Jiaoyang Li, Dor Atzmon, Liron Cohen, TK~Kumar, et~al.
\newblock Multi-agent pathfinding: Definitions, variants, and benchmarks.
\newblock In \emph{Proceedings of the International Symposium on Combinatorial Search}, volume~10, pages 151--158, 2019.

\bibitem[Li et~al.(2021{\natexlab{a}})Li, Tinka, Kiesel, Durham, Kumar, and Koenig]{li2021lifelong}
Jiaoyang Li, Andrew Tinka, Scott Kiesel, Joseph~W Durham, TK~Satish Kumar, and Sven Koenig.
\newblock Lifelong multi-agent path finding in large-scale warehouses.
\newblock In \emph{Proceedings of the AAAI Conference on Artificial Intelligence}, volume~35, pages 11272--11281, 2021{\natexlab{a}}.

\bibitem[Hopcroft et~al.(1984)Hopcroft, Schwartz, and Sharir]{hopcroft1984complexity}
John~E Hopcroft, Jacob~Theodore Schwartz, and Micha Sharir.
\newblock On the complexity of motion planning for multiple independent objects; pspace-hardness of the" warehouseman's problem".
\newblock \emph{The international journal of robotics research}, 3\penalty0 (4):\penalty0 76--88, 1984.

\bibitem[Song and Ermon(2019)]{song2019generative}
Yang Song and Stefano Ermon.
\newblock Generative modeling by estimating gradients of the data distribution.
\newblock \emph{Advances in neural information processing systems}, 32, 2019.

\bibitem[Ho et~al.(2020)Ho, Jain, and Abbeel]{ho2020denoising}
Jonathan Ho, Ajay Jain, and Pieter Abbeel.
\newblock Denoising diffusion probabilistic models.
\newblock \emph{Advances in neural information processing systems}, 33:\penalty0 6840--6851, 2020.

\bibitem[Carvalho et~al.(2023)Carvalho, Le, Baierl, Koert, and Peters]{carvalho2023motion}
Joao Carvalho, An~T Le, Mark Baierl, Dorothea Koert, and Jan Peters.
\newblock Motion planning diffusion: Learning and planning of robot motions with diffusion models.
\newblock In \emph{2023 IEEE/RSJ International Conference on Intelligent Robots and Systems (IROS)}, pages 1916--1923. IEEE, 2023.

\bibitem[Christopher et~al.(2024)Christopher, Baek, and Fioretto]{christopher2024constrained}
Jacob~K Christopher, Stephen Baek, and Ferdinando Fioretto.
\newblock Constrained synthesis with projected diffusion models.
\newblock In \emph{The Thirty-eighth Annual Conference on Neural Information Processing Systems}, 2024.

\bibitem[Li et~al.(2019)Li, Felner, Boyarski, Ma, and Koenig]{li2019improved}
Jiaoyang Li, Ariel Felner, Eli Boyarski, Hang Ma, and Sven Koenig.
\newblock Improved heuristics for multi-agent path finding with conflict-based search.
\newblock In \emph{IJCAI}, volume 2019, pages 442--449, 2019.

\bibitem[Li et~al.(2021{\natexlab{b}})Li, Ruml, and Koenig]{li2021eecbs}
Jiaoyang Li, Wheeler Ruml, and Sven Koenig.
\newblock Eecbs: A bounded-suboptimal search for multi-agent path finding.
\newblock In \emph{Proceedings of the AAAI conference on artificial intelligence}, volume~35, pages 12353--12362, 2021{\natexlab{b}}.

\bibitem[Okumura et~al.(2022{\natexlab{a}})Okumura, Machida, D{\'e}fago, and Tamura]{okumura2022priority}
Keisuke Okumura, Manao Machida, Xavier D{\'e}fago, and Yasumasa Tamura.
\newblock Priority inheritance with backtracking for iterative multi-agent path finding.
\newblock \emph{Artificial Intelligence}, 310:\penalty0 103752, 2022{\natexlab{a}}.

\bibitem[Shaoul et~al.(2024)Shaoul, Mishani, Vats, Li, and Likhachev]{shaoul2024multi}
Yorai Shaoul, Itamar Mishani, Shivam Vats, Jiaoyang Li, and Maxim Likhachev.
\newblock Multi-robot motion planning with diffusion models.
\newblock \emph{arXiv preprint arXiv:2410.03072}, 2024.

\bibitem[Kavraki et~al.(1996)Kavraki, Svestka, Latombe, and Overmars]{kavraki1996probabilistic}
Lydia~E Kavraki, Petr Svestka, J-C Latombe, and Mark~H Overmars.
\newblock Probabilistic roadmaps for path planning in high-dimensional configuration spaces.
\newblock \emph{IEEE transactions on Robotics and Automation}, 12\penalty0 (4):\penalty0 566--580, 1996.

\bibitem[LaValle(1998)]{lavalle1998rapidly}
Steven LaValle.
\newblock Rapidly-exploring random trees: A new tool for path planning.
\newblock \emph{Research Report 9811}, 1998.

\bibitem[Augugliaro et~al.(2012)Augugliaro, Schoellig, and D'Andrea]{6385823}
Federico Augugliaro, Angela~P. Schoellig, and Raffaello D'Andrea.
\newblock Generation of collision-free trajectories for a quadrocopter fleet: A sequential convex programming approach.
\newblock In \emph{2012 IEEE/RSJ International Conference on Intelligent Robots and Systems}, pages 1917--1922, 2012.
\newblock \doi{10.1109/IROS.2012.6385823}.

\bibitem[Chen et~al.(2015)Chen, Cutler, and How]{7140034}
Yufan Chen, Mark Cutler, and Jonathan~P. How.
\newblock Decoupled multiagent path planning via incremental sequential convex programming.
\newblock In \emph{2015 IEEE International Conference on Robotics and Automation (ICRA)}, pages 5954--5961, 2015.
\newblock \doi{10.1109/ICRA.2015.7140034}.

\bibitem[Chen et~al.(2023)Chen, Liang, Cheng, You, and Yang]{10286111}
Ruishuang Chen, Zhihui Liang, Jie Cheng, Pengcheng You, and Zaiyue Yang.
\newblock Multi-agent cooperative motion planning based on alternating direction method of multipliers.
\newblock \emph{IEEE Control Systems Letters}, 7:\penalty0 3307--3312, 2023.
\newblock \doi{10.1109/LCSYS.2023.3324663}.

\bibitem[Janner et~al.(2022)Janner, Du, Tenenbaum, and Levine]{janner2022planning}
Michael Janner, Yilun Du, Joshua~B Tenenbaum, and Sergey Levine.
\newblock Planning with diffusion for flexible behavior synthesis.
\newblock \emph{arXiv preprint arXiv:2205.09991}, 2022.

\bibitem[Okumura et~al.(2022{\natexlab{b}})Okumura, Yonetani, Nishimura, and Kanezaki]{okumura2022ctrms}
Keisuke Okumura, Ryo Yonetani, Mai Nishimura, and Asako Kanezaki.
\newblock Ctrms: Learning to construct cooperative timed roadmaps for multi-agent path planning in continuous spaces.
\newblock \emph{arXiv preprint arXiv:2201.09467}, 2022{\natexlab{b}}.

\bibitem[Yang et~al.(2023)Yang, Zhang, Song, Hong, Xu, Zhao, Zhang, Cui, and Yang]{yang2023diffusion}
Ling Yang, Zhilong Zhang, Yang Song, Shenda Hong, Runsheng Xu, Yue Zhao, Wentao Zhang, Bin Cui, and Ming-Hsuan Yang.
\newblock Diffusion models: A comprehensive survey of methods and applications.
\newblock \emph{ACM Computing Surveys}, 56\penalty0 (4):\penalty0 1--39, 2023.

\bibitem[Song et~al.(2020)Song, Sohl-Dickstein, Kingma, Kumar, Ermon, and Poole]{song2020score}
Yang Song, Jascha Sohl-Dickstein, Diederik~P Kingma, Abhishek Kumar, Stefano Ermon, and Ben Poole.
\newblock Score-based generative modeling through stochastic differential equations.
\newblock \emph{arXiv preprint arXiv:2011.13456}, 2020.

\bibitem[Boyd et~al.(2011)Boyd, Parikh, Chu, Peleato, Eckstein, et~al.]{boyd2011distributed}
Stephen Boyd, Neal Parikh, Eric Chu, Borja Peleato, Jonathan Eckstein, et~al.
\newblock Distributed optimization and statistical learning via the alternating direction method of multipliers.
\newblock \emph{Foundations and Trends{\textregistered} in Machine learning}, 3\penalty0 (1):\penalty0 1--122, 2011.

\bibitem[Kotary et~al.(2022)Kotary, Fioretto, and Van~Hentenryck]{kotary2022fast}
James Kotary, Ferdinando Fioretto, and Pascal Van~Hentenryck.
\newblock Fast approximations for job shop scheduling: A lagrangian dual deep learning method.
\newblock In \emph{Proceedings of the AAAI Conference on Artificial Intelligence}, volume~36, pages 7239--7246, 2022.

\end{thebibliography}

\end{document}